\documentclass[conference]{IEEEtran}
\IEEEoverridecommandlockouts
\usepackage{hyperref}
\usepackage{cite}
\usepackage{multirow}
\usepackage{amsmath,amssymb,amsfonts}
\usepackage{algorithmic}
\usepackage{graphicx}
\usepackage{textcomp}
\usepackage{mathtools}
\usepackage{multicol}
\usepackage{xcolor}
\usepackage{colortbl}
\usepackage{comment}

\usepackage[numbers,sort&compress]{natbib}
\usepackage{booktabs}
\usepackage{siunitx}
\usepackage[accsupp]{axessibility}  
\usepackage{soul}
\usepackage{bm}
\usepackage{subcaption}

\usepackage{pgf}
\usepackage{tikz}
\usepackage{tikzscale}
\usetikzlibrary{arrows,external}
\usetikzlibrary{spy}
\usepackage{pgfplots}
\usepackage[skins]{tcolorbox}
\usepackage{import}
\usepackage{url}
\usepackage{tabu} 

\newcommand{\signtok}{\textsc{SignTok}}

\def\BibTeX{{\rm B\kern-.05em{\sc i\kern-.025em b}\kern-.08em
    T\kern-.1667em\lower.7ex\hbox{E}\kern-.125emX}}

\title{\signtok: A Text-Aligned Continuous Pose Tokenizer for Sign Language Generation}

\author{Nabeela Khan$^{1,2,*}$, Gaurav Rai$^{3,*}$, Bowen Wu$^{2,4}$, Sihan Tan$^{1}$, Benjamin Yen$^{5,1}$, \\ Takeshi Ashizawa$^{1}$, and Kazuhiro Nakadai$^{1}$
\thanks{$^{*}$These authors contributed equally to this work.}%
\thanks{This work was supported by KAKENHI 26H02525, and ROIS NII Open Collaborative Research 261S04-24183.}
\thanks{$^{1}$Department of Systems and Control Engineering, Institute of Science Tokyo, Tokyo, Japan
        {\tt\small nabeela@ra.sc.e.titech.ac.jp}}%
\thanks{$^{2}$RIKEN Guardian Robot Project, Kyoto, Japan}%
\thanks{$^{3}$IIIT Delhi, India
        {\tt\small gauravr@iiitd.ac.in}}%
\thanks{$^{4}$ATR Hiroshi Ishiguro Laboratories, Kyoto, Japan}%
\thanks{$^{5}$RIKEN Center for Biosystems Dynamics Research, Kobe, Japan}%
}

\begin{document}

\maketitle

\begin{abstract}
Text-conditioned sign language generation (SLG) depends not only on the generator but also on the space in which signing is generated. Existing approaches generate directly in pose space, in quantized code spaces, or in learned latent spaces trained for reconstruction. We find that these spaces are organized largely by signer identity rather than linguistic content: on CSL-Daily, a test clip's nearest neighbor in raw pose space is the same sentence signed by another person in only 2\% of cases. To address this, we propose \signtok, a continuous pose tokenizer whose latent space is shaped by three training-only heads: a kinematics head that predicts motion dynamics, a sentence-level head that aligns signing with the overall text, and a word-level head that aligns short motion windows with content words. Together, these objectives raise the same-sentence cross-signer nearest-neighbor rate to 82\%. On three sign language benchmarks, the same generator trained in the \signtok\ latent instead of pose space raises back-translation BLEU-4 in five of six benchmark-scorer settings at comparable joint accuracy, and word-level alignment gives higher back-translation scores than sentence-level alignment on every benchmark. Code will be made available soon.
\end{abstract}

\section{Introduction}
\label{sec:intro}

\begin{figure*}[t]
\centering
\includegraphics[width=0.9\textwidth]{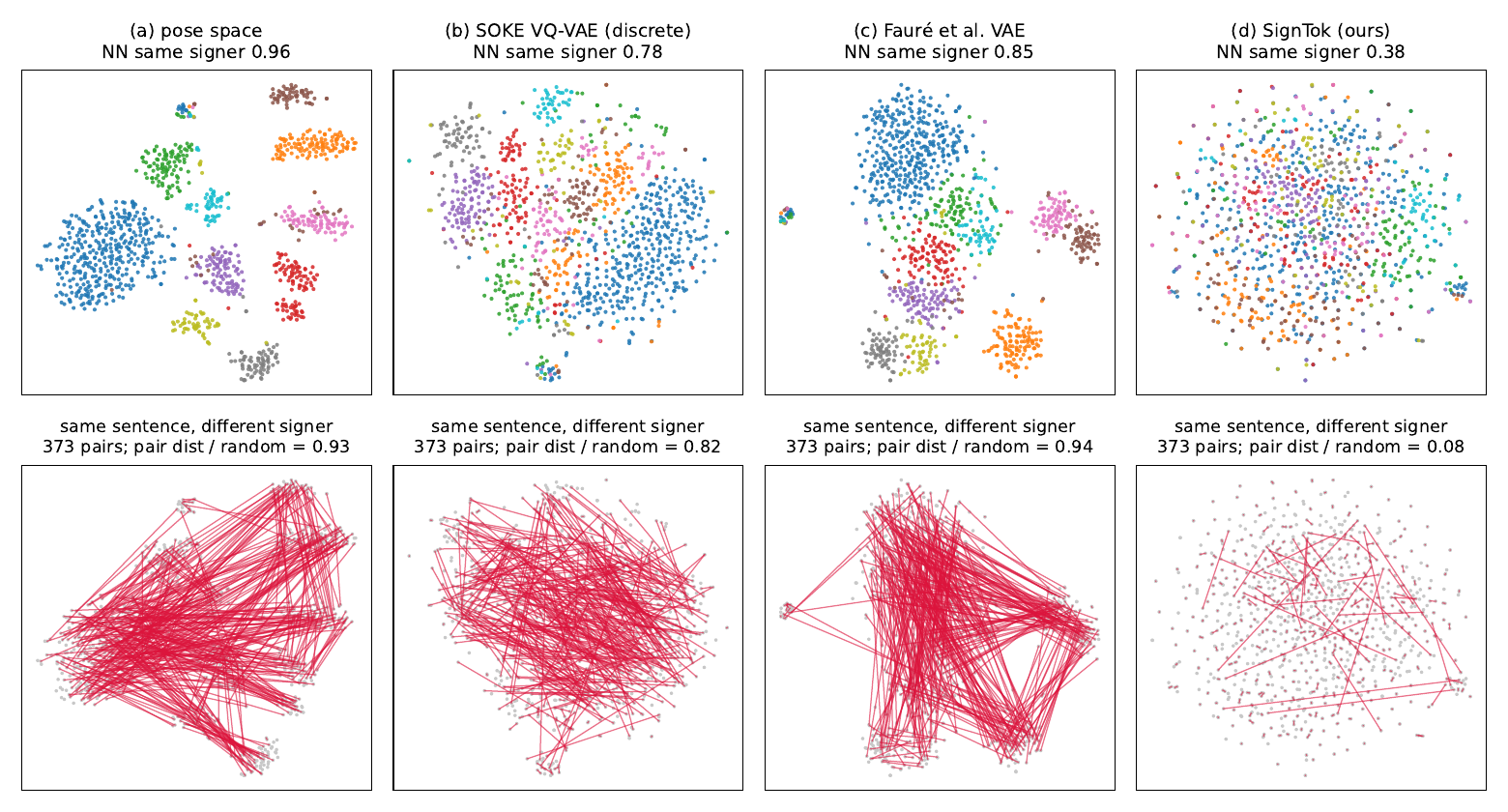}
\caption{\textbf{What organizes the space a sign generator works in?} t-SNE of mean-pooled CSL-Daily test clips, coloured by signer (ten most frequent), for (a) the pose space itself, (b) the discrete tokenizer of SOKE~\cite{zuo2025signs}, (c) the sign pose VAE of Faur\'e et al.~\cite{faure2026impact}, both trained with the released code on the same poses, and (d) \signtok. In (a)-(c) every signer forms clusters: the representations that current generators map text into are organized by \emph{who} signed, not by \emph{what} was signed.}
\label{fig:related-latent}
\end{figure*}

End-to-end sign language generation (SLG) converts spoken text into sign language (SL) motion, making language technologies more inclusive for the Deaf and Hard-of-Hearing communities~\cite{hakozaki2025evaluation,khan2026sign}. SL is a demanding target: high-dimensional manual and non-manual pose components, transitions between signs~\cite{low2026signspark}, signer-specific style, and a one-to-many mapping from text~\cite{khan2025signflow}. Much of this variation is not linguistic: translation scores drop sharply under signer-independent splits~\cite{artiaga2025rethinking}, and in pose space the same sentence signed by two people is almost as far apart as two unrelated sentences.

Recent SLG approaches use deterministic sequence generation~\cite{saunders2020progressive, khan2025multigau, hwang2021non}, probabilistic generation~\cite{khan2025signflow, low2026signspark, baltatzis2024neural, rastgoo2025diffusion}, and language models over quantized motion codes~\cite{yin2024t2s, zuo2025signs, walsh2024data, zuo2026madis}. Two generation spaces dominate. \emph{Discrete} tokenizers quantize poses into codebook entries~\cite{yin2024t2s, zuo2025signs, walsh2024data, guo2024momask}, and fine hand detail can be lost in the quantization. \emph{Continuous} generators synthesize poses directly with diffusion or flow matching~\cite{baltatzis2024neural, khan2025signflow, low2026signspark}; they keep precision but must denoise a pose space in which much of the variation is not linguistic. A continuous tokenizer sits between the two: it compresses the pose into a dense per-frame latent that can be generated without quantization loss, as established for human motion~\cite{chen2023executing, meng2025rethinking} and recently explored for SL~\cite{faure2026impact, hwang2021non}. Work on image generation has shown that such a latent should be shaped for the generator, not for reconstruction alone~\cite{yao2025reconstruction, chen2025masked, yu2024representation}. This raises our question: \emph{how should the latent of a sign tokenizer be organized}, when SL mixes linguistic content with signer style and articulation detail that the text does not determine?

Since SLG is conditioned on text, the generator must map each sentence into the latent, and a space organized mainly by the signer gives it little help. Prior work points the same way: semantic and language guidance make image latents easier to model~\cite{zha2025language, yao2025reconstruction}; MotionCLIP~\cite{tevet2022motionclip} and SignCLIP~\cite{jiang2024signclip} align motion and signing with text at the clip level; SeRV~\cite{wu2026serv} aligns a discrete sign tokenizer with the contextual states of a text encoder, a target that our analysis shows favours sentence identity over individual words; and SignRep~\cite{wong2025signrep} suppresses signer appearance adversarially, but for recognition rather than generation, and removing style does not by itself tie the latent to the text. Inspired by these works, we hypothesize that what is missing here is a continuous latent, built for generation, whose text alignment is fine enough to follow individual signs rather than whole clips.

We propose \signtok, a continuous pose tokenizer: a masked autoencoder over frames trained with three shallow auxiliary heads. A kinematics head predicts the velocity and acceleration of the masked frames, so the latent encodes how the signer moves through the masked frames; a sentence head regresses the pooled latent onto the sentence feature of a frozen text encoder, pulling the clips of one sentence together; and a word-level alignment head matches short windows of the latent to the content words of the sentence through a late-interaction contrastive objective, so that every word is realized by some stretch of motion and the same word lands in the same region whoever signs it. Together, they organize the latent by what is signed, without removing who signs it.

Our contributions are as follows:
\begin{enumerate}\setlength\itemsep{0pt}
\item We show that pose space and sign pose tokenizers trained for reconstruction are organized by signer identity, with limited semantic organization: on CSL-Daily, the nearest neighbor of a clip is the same sentence by another signer in at most 9\% of cases.
\item We propose \signtok, a masked pose autoencoder whose latent space is shaped by three training-only auxiliary heads for kinematic modeling, sentence-level alignment, and word-level alignment, which raises that rate to 82\%, and we show that the word-level head localizes words rather than recognizing sentences.
\item On three standard SLG benchmarks, the same generator trained in the \signtok\ latent instead of pose space raises back-translation BLEU-4 in five of six benchmark-scorer settings at comparable joint accuracy.
\end{enumerate}

\begin{figure*}[t]
\centering
\includegraphics[width=0.9\textwidth]{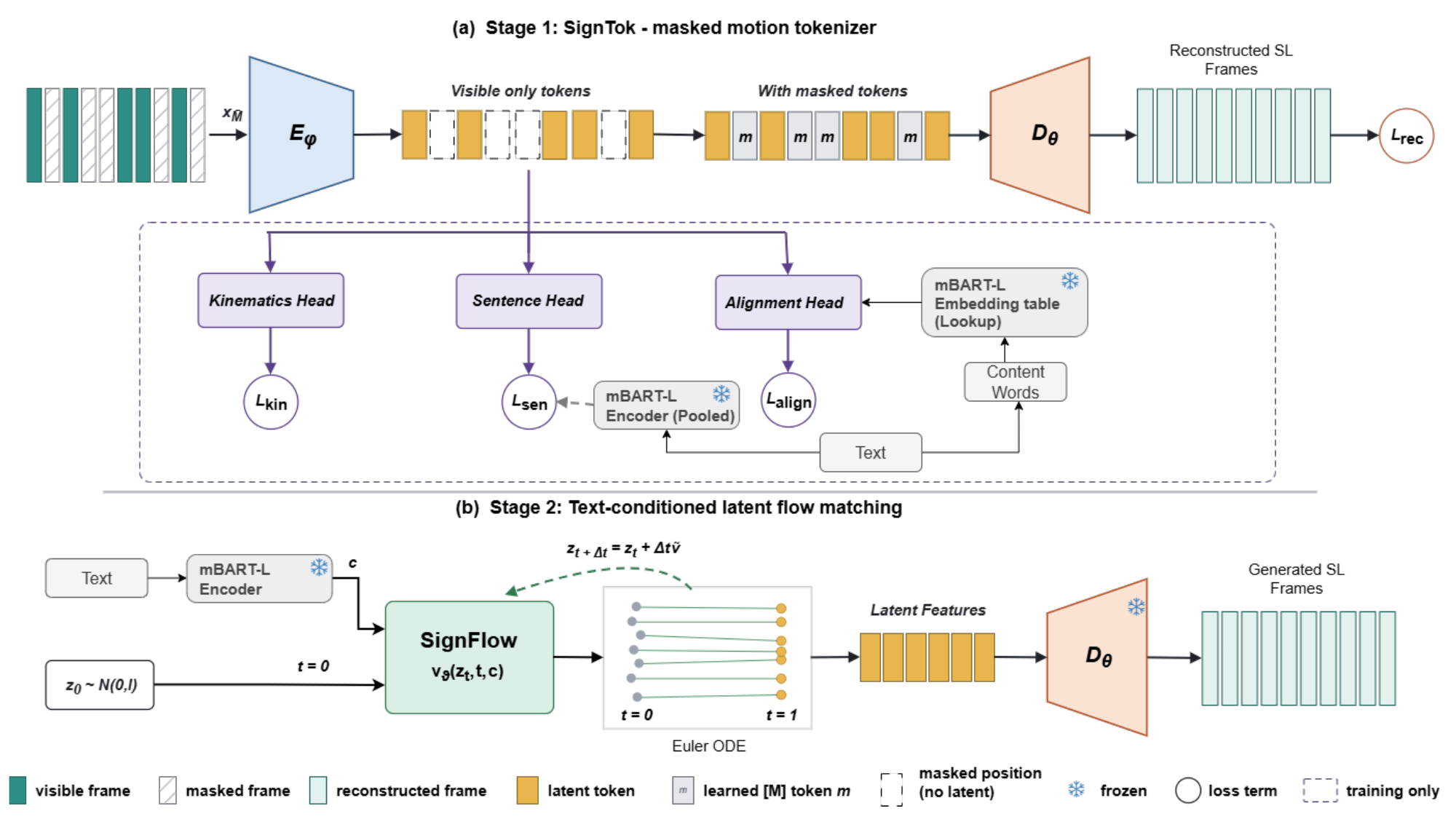}
\caption{\textbf{\signtok\ overview.} (a) Stage 1: the encoder $E_\varphi$ sees only the visible frames of a masked clip and yields one latent per visible frame; the decoder input is the latent completed with a learned token $m$ at the masked positions, and $D_\theta$ reconstructs the clip. Three training-only heads read the latent: kinematics, sentence, and word-level alignment. (b) Stage 2: with the tokenizer frozen and no masking, a flow-matching model transports noise to the per-frame latents under cross-attention to the frozen mBART states of the sentence, and the frozen decoder gives the poses.}
\label{fig:arch}
\end{figure*}

\section{Related Work}
\label{sec:related}

\subsection{Representations for sign language generation}
SLG has been approached with deterministic sequence models~\cite{saunders2020progressive, khan2025multigau, hwang2021non}, probabilistic generators~\cite{khan2025signflow, low2026signspark, baltatzis2024neural, rastgoo2025diffusion} and language models over motion codes~\cite{yin2024t2s, zuo2025signs, walsh2024data, zuo2026madis}; what they share is an intermediate representation between text and poses. \textit{Glosses}\footnote{A gloss is a label for an isolated sign, often used as an intermediate symbolic representation.} are the oldest: Progressive Transformers \cite{saunders2020progressive} and their successors \cite{saunders2020adversarial, saunders2020everybody} translate text to glosses and regress poses from them, and gloss-to-pose diffusion keeps that interface \cite{tang2025sign, xie2024g2p}; HamNoSys \cite{uchida2024hamnosys}, motion vocabularies \cite{walsh2024data} and dictionary retrieval with stitching \cite{walsh2024sign, fang2026stable} are symbolic alternatives with the same costs, expensive annotation and an information bottleneck. Generating directly in pose space, by regression \cite{saunders2020progressive}, diffusion \cite{baltatzis2024neural, feng2025text2sign} or flow matching \cite{khan2025signflow, low2026signspark}, removes the symbols but leaves the generator in a high-dimensional, strongly signer-dependent space. Learned spaces compress the poses first: discrete codebooks emitted from text by autoregressive or discrete-diffusion models \cite{hwang2023autoregressive, yin2024t2s, guo2024momask}, including SOKE \cite{zuo2025signs} with one VQ-VAE per articulator and SeRV \cite{wu2026serv} with a residual codebook and a text-alignment loss; and continuous latents following latent diffusion for human motion \cite{chen2023executing, meng2025rethinking}, with Faur\'e et al.~\cite{faure2026impact} comparing VAE designs and A$^2$V-SLP \cite{tacsyurek20262} imposing articulator-wise latents. With the exception of SeRV, the representation is learned to reconstruct the poses, and the text enters only afterwards, through the generator. Figure~\ref{fig:related-latent} shows what this produces for the pose space (a), SOKE's latent (b) and the VAE of Faur\'e~\cite{faure2026impact} (c): clips cluster by signer, and two renditions of a sentence lie almost as far apart as two random clips. The same holds on PHOENIX-2014T and How2Sign (see the supplementary material).

\begin{figure*}[t]
\centering
\includegraphics[width=0.9\textwidth]{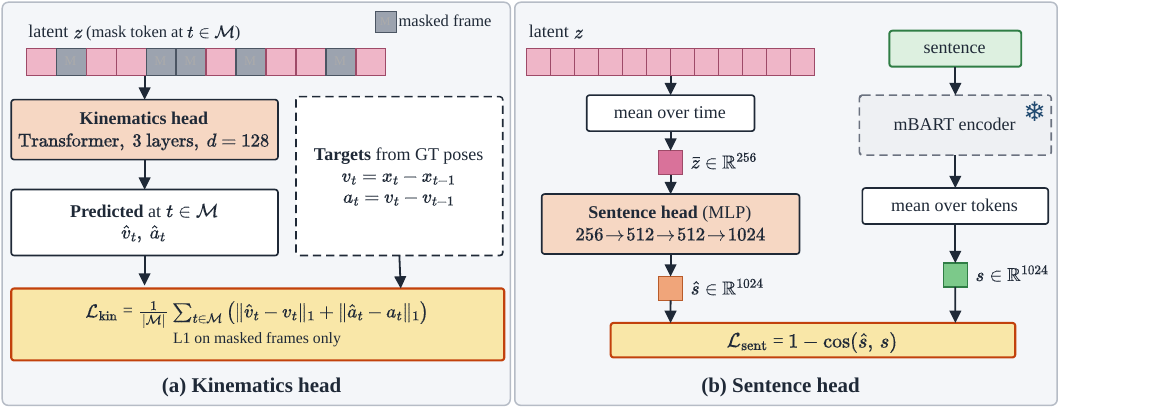}
\caption{\textbf{Kinematics and sentence heads.} (a) A shallow Transformer over the latent, with its own mask token at the masked positions, predicts the velocity and acceleration of the masked frames from finite differences of the true poses. (b) An MLP maps the time-pooled latent to the token-pooled state of the frozen mBART encoder under a cosine loss.}
\label{fig:heads}
\end{figure*}

\subsection{Text-aligned representation learning for generation}
Outside SLG, what a generator's latent should encode has been separated from how well it reconstructs: VA-VAE \cite{yao2025reconstruction} aligns the latent with a vision foundation model, REPA and REPA-E \cite{yu2024representation, leng2025repa} align diffusion features and end-to-end tuned VAEs with DINOv2, MAETok \cite{chen2025masked} reads the latent of a masked autoencoder through shallow auxiliary decoders discarded at inference, and AlignTok, SVG, a study of ``diffusable'' latents and work on motion tokenizers reach similar conclusions \cite{chen2025aligntok, shi2026latent, zhong2026diffusing, yan2026beyond, meng2025absolute}. The alignment targets in this line are \emph{condition-blind}: a self-supervised encoder of the same modality, with language used at most as guidance on the generator side \cite{zha2025language}. Alignment with text exists, but mostly for retrieval or translation rather than as a generator's working space: CLIP-style losses tie a clip to a sentence \cite{radford2021learning, tevet2022motionclip, jiang2024signclip, zhou2023gloss}, late-interaction losses tie tokens to parts \cite{yao2021filip, khattab2020colbert, zhang2026fine, hu2026vtamo}, SeRV \cite{wu2026serv} brings both into a discrete sign tokenizer for generation, with the contextual states of a text encoder as the target, and SignRep \cite{wong2025signrep} suppresses signer appearance adversarially for recognition, which removes style without tying the latent to the text. To our knowledge, none of them shapes a continuous latent that a text-conditioned generator must reach by the individual words it will be conditioned on. \signtok\ takes MAETok's auxiliary-decoder recipe and makes the condition itself, the words of the sentence, the alignment target of a continuous latent for flow matching.

\section{\signtok}
\label{sec:method}

Figure~\ref{fig:arch} gives an overview of the approach, which has two stages. In the first, a masked autoencoder over pose frames learns a per-frame latent (Section~\ref{sec:tokenizer}), shaped by three training-only heads for kinematics, sentence-level alignment and word-level alignment (Section~\ref{sec:heads}). In the second, the tokenizer is frozen and a text-conditioned flow-matching generator produces the latents, which the frozen decoder renders to poses (Section~\ref{sec:signflow}).

\subsection{Masked tokenizer}
\label{sec:tokenizer}

The task of the masked tokenizer is to learn a representation of the sign sequence that is better suited to generation than the poses themselves. Its input is a sign sequence $x=x_{1:T}$ with $x_t\in\mathbb{R}^{D}$, where $T$ is the number of frames and $D$ the dimension of the per-frame pose feature. For every sequence, a random set $M$ of $\lfloor rT\rfloor$ frames is masked, with $r$ the masking ratio; $\bar M$ denotes the visible frames. Following MAE \cite{he2022masked}, the masked frames are dropped before the encoder, which sees the visible frames only, and mask tokens enter at the decoder.

\begin{equation}
z=E_\varphi(x_{\bar M}),\qquad z_t\in\mathbb{R}^{d},\ t\in\bar M,
\end{equation}
The decoder input $\tilde z\in\mathbb{R}^{T\times d'}$ is the latent projected to the decoder width and completed with one learned token $m$ at every masked position,
\begin{equation}
\tilde z_t=\begin{cases} W z_t & t\in\bar M\\ m & t\in M,\end{cases}
\qquad \hat x=D_\theta(\tilde z),
\end{equation}
so the decoder must infer each masked frame from its visible neighbours. The reconstruction loss is an L1 on the masked frames with a small weight $\lambda_{\text{v}}$ on the visible ones,
\begin{equation}
\begin{aligned}
\mathcal{L}_{\text{rec}}=\tfrac{1}{|M|}\textstyle\sum_{t\in M}&\|\hat x_t-x_t\|_1\\[-2pt]
+\lambda_{\text{v}}\tfrac{1}{|\bar M|}\textstyle\sum_{t\in\bar M}&\|\hat x_t-x_t\|_1 .
\end{aligned}
\end{equation}

\subsection{Auxiliary latent losses}
\label{sec:heads}

\paragraph{Kinematics head (Figure~\ref{fig:heads}a).}
A shallow Transformer $h_{\text{kin}}$ takes the latent completed with its own mask token $m'$ and predicts, for the masked frames, the velocity $v_t=x_t-x_{t-1}$ and acceleration $a_t=x_{t+1}-2x_t+x_{t-1}$ of the true poses,
\begin{equation}
\mathcal{L}_{\text{kin}}=\tfrac{1}{|M|}\textstyle\sum_{t\in M}\big\|[\hat v_t,\hat a_t]-[v_t,a_t]\big\|_2^2 .
\end{equation}

\paragraph{Sentence head (Figure~\ref{fig:heads}b).}
With $s$ the mean of the frozen mBART encoder states over the tokens of the sentence, an MLP $h_{\text{sent}}$ regresses it from the latent pooled over the visible frames,
\begin{equation}
\hat s=h_{\text{sent}}\big(\tfrac{1}{|\bar M|}\textstyle\sum_{t\in\bar M}z_t\big),\quad
\mathcal{L}_{\text{sent}}=1-\cos(\hat s,s).
\end{equation}

\begin{figure*}[t]
\centering
\includegraphics[width=\textwidth]{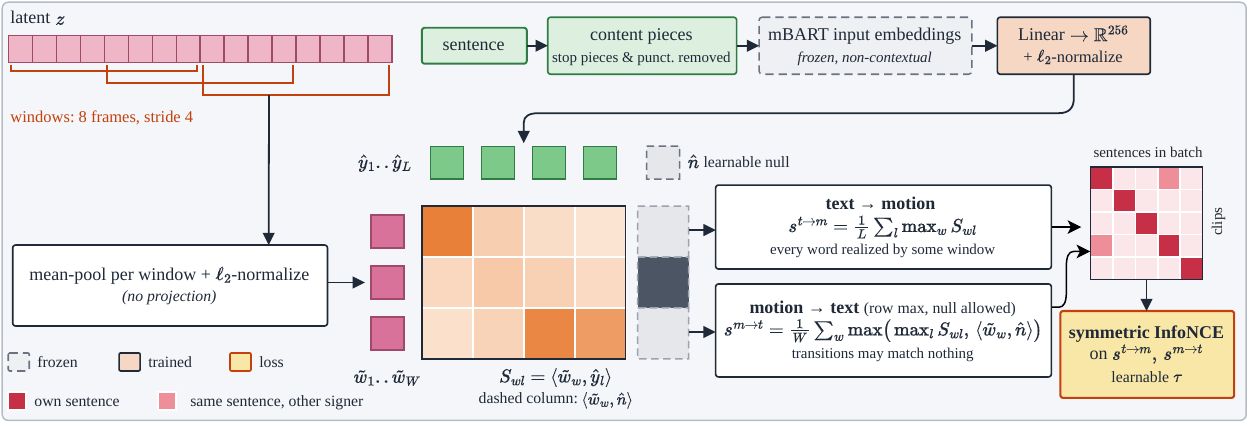}
\caption{\textbf{Word-level alignment head.} Non-overlapping windows of the latent are mean-pooled and $\ell_2$-normalized without projection. The content pieces of the sentence are looked up in the frozen input embedding table of mBART and projected into the latent space, together with a learnable null vector $n$. Two late-interaction scores (best window per token; best token or null per window) enter a symmetric InfoNCE over the batch in which the same sentence signed by another person is also a positive.}
\label{fig:alignhead}
\end{figure*}

\paragraph{Word-level alignment head (Figure~\ref{fig:alignhead}).}
This head aligns \emph{segments} of a sequence to \emph{words} with a late-interaction objective in the style of FILIP \cite{yao2021filip}. The latent is cut into non-overlapping windows of $k$ frames; $\hat w_w$ is the $\ell_2$-normalized mean of the latents of the visible frames in window $w$, with no projection. On the text side, $y_{1:L}$ are the rows of mBART's \emph{input} embedding table for the content pieces of the sentence (pieces that are not special, not punctuation and not in a stop list of frequent pieces), so a token carries its word and nothing about the sentence around it. A linear map $P$ brings them into the latent space, together with a learnable null vector $n$:
\begin{equation}
\hat y_l=\tfrac{P y_l}{\|P y_l\|},\qquad \hat n=\tfrac{n}{\|n\|},\qquad S_{wl}=\langle\hat w_w,\hat y_l\rangle .
\end{equation}
For SL sequence $i$ and sentence $j$ in the batch,
\begin{align}
s^{\text{t}\to\text{m}}_{ij} &= \tfrac{1}{L_j}\textstyle\sum_{l}\max_{w} S^{ij}_{wl},\label{eq:t2m}\\
s^{\text{m}\to\text{t}}_{ij} &= \tfrac{1}{W_i}\textstyle\sum_{w}\max\big(\max_{l} S^{ij}_{wl},\,\langle\hat w_w,\hat n\rangle\big).\label{eq:m2t}
\end{align}
Eq.~\ref{eq:t2m} takes the best window for every token, so every word must be realized somewhere in the SL sequence; Eq.~\ref{eq:m2t} takes the best token for every window but lets a window match the null instead, so transitions between signs need not match any word, and since $\langle\hat w_w,\hat n\rangle$ does not depend on $j$, such a window stops discriminating between sentences. Each score enters a symmetric InfoNCE over the batch with learnable temperature $\tau$; because sentences recur, signed by different people, the target is multi-positive. With $\mathcal{P}_i$ the batch items whose sentence equals that of seq $i$,
\begin{equation}
\begin{aligned}
\mathcal{L}_{\text{NCE}}(s)=-\tfrac{1}{2B}\textstyle\sum_{i}\tfrac{1}{|\mathcal{P}_i|}\sum_{j\in\mathcal{P}_i}\Big[&\log\tfrac{e^{s_{ij}/\tau}}{\sum_{j'}e^{s_{ij'}/\tau}}\\[-2pt]
+\,&\log\tfrac{e^{s_{ji}/\tau}}{\sum_{i'}e^{s_{ji'}/\tau}}\Big],
\end{aligned}
\end{equation}
\begin{equation}
\mathcal{L}_{\text{align}}=\tfrac{1}{1+\beta}\big(\mathcal{L}_{\text{NCE}}(s^{\text{t}\to\text{m}})+\beta\,\mathcal{L}_{\text{NCE}}(s^{\text{m}\to\text{t}})\big).
\end{equation}
Three choices distinguish this head from FILIP: (i)~the text side is non-contextual, so matching a token means matching a word, not the sentence it occurs in; (ii)~the null target, without which Eq.~\ref{eq:m2t} forces every window, transitions included, onto some word; (iii)~the motion side has no projection, so what the head learns is imposed on the latent the generator will see.

The three heads read the latent during training only and are discarded afterwards; the full objective is
\begin{equation}
\mathcal{L}=\mathcal{L}_{\text{rec}}+\lambda_{\text{kin}}\mathcal{L}_{\text{kin}}+\lambda_{\text{sent}}\mathcal{L}_{\text{sent}}+\lambda_{\text{align}}\mathcal{L}_{\text{align}}.
\end{equation}

\subsection{Generation in the latent}
\label{sec:signflow}
The generator is SignFlow \cite{khan2025signflow}, a text-conditioned flow-matching model, used unchanged except for the space it generates in: instead of poses, it generates the per-frame latents of the frozen tokenizer, which the frozen decoder renders to poses. Architecture, optimal-transport pairing, and guidance follow SignFlow~\cite{khan2025signflow}.

\definecolor{grphead}{RGB}{221,232,246}   
\definecolor{ourrow}{RGB}{240,246,252}    
\definecolor{refrow}{RGB}{242,242,242}    
\newcommand{\snd}[1]{\underline{#1}}      

\begin{table*}[t]
\centering\small
\setlength{\tabcolsep}{2.0pt}
\resizebox{\textwidth}{!}{%
\begin{tabular}{l cccccc cccccc cccccc}
\toprule
& \multicolumn{6}{c}{\textbf{CSL-Daily}} & \multicolumn{6}{c}{\textbf{PHOENIX-2014T}} & \multicolumn{6}{c}{\textbf{How2Sign}} \\
\cmidrule(lr){2-7}\cmidrule(lr){8-13}\cmidrule(lr){14-19}
& \multicolumn{2}{c}{PA-JPE$\downarrow$} & \multicolumn{2}{c}{JPE$\downarrow$} & \multicolumn{2}{c}{B-T$\uparrow$}
& \multicolumn{2}{c}{PA-JPE$\downarrow$} & \multicolumn{2}{c}{JPE$\downarrow$} & \multicolumn{2}{c}{B-T$\uparrow$}
& \multicolumn{2}{c}{PA-JPE$\downarrow$} & \multicolumn{2}{c}{JPE$\downarrow$} & \multicolumn{2}{c}{B-T$\uparrow$} \\
\cmidrule(lr){2-3}\cmidrule(lr){4-5}\cmidrule(lr){6-7}\cmidrule(lr){8-9}\cmidrule(lr){10-11}\cmidrule(lr){12-13}\cmidrule(lr){14-15}\cmidrule(lr){16-17}\cmidrule(lr){18-19}
Method & Body & Hand & Body & Hand & S & U & Body & Hand & Body & Hand & S & U & Body & Hand & Body & Hand & S & U \\
\midrule
\rowcolor{grphead}\multicolumn{19}{c}{\textbf{Gloss-free text-to-pose, as published}} \\
Prog.\ Transformer$^\dagger$ & 15.98 & 12.91 & 16.30 & 32.63 & -- & -- & 13.67 & 11.95 & 15.01 & 31.77 & -- & -- & 14.15 & 11.57 & 14.74 & 30.17 & -- & -- \\
Text2Mesh$^\dagger$ & 13.47 & 12.10 & 13.76 & 30.37 & -- & -- & 13.48 & 12.06 & 14.04 & 31.64 & -- & -- & 13.99 & 13.47 & 15.50 & 32.97 & -- & -- \\
T2S-GPT$^\dagger$ & 11.94 & 5.93 & 12.32 & 15.43 & -- & -- & 10.38 & 6.47 & 11.65 & 19.09 & -- & -- & 11.48 & 6.39 & 12.65 & 18.44 & -- & -- \\
S-MotionGPT$^\dagger$ & 10.81 & 3.78 & 11.58 & 11.31 & -- & -- & 9.45 & 3.41 & 10.42 & 9.08 & -- & -- & 11.23 & 4.39 & 12.41 & 13.74 & -- & -- \\
SOKE & 7.58 & 2.17 & -- & -- & -- & -- & 6.16 & 1.85 & -- & -- & -- & -- & 7.91 & 3.10 & -- & -- & -- & -- \\
SignSparK & 7.26 & 2.00 & 6.27 & 10.63 & -- & -- & 5.24 & 1.52 & 4.40 & 7.10 & -- & -- & 7.26 & 2.72 & 6.30 & 11.43 & -- & -- \\
\rowcolor{grphead}\multicolumn{19}{c}{\textbf{With gloss-based sign retrieval, as published} \emph{(not comparable to text-only rows)}} \\
SOKE & 6.24 & 1.71 & 7.38 & 9.68 & -- & -- & 4.77 & 1.38 & 6.04 & 7.72 & -- & -- & 6.82 & 2.35 & 7.75 & 10.08 & -- & -- \\
SignSparK & 4.87 & 1.37 & 4.89 & 6.56 & -- & -- & 3.68 & 1.18 & 3.51 & 4.82 & -- & -- & -- & -- & -- & -- & -- & -- \\
\rowcolor{grphead}\multicolumn{19}{c}{\textbf{Gloss-free text-to-pose on the released poses} \emph{(same clips, scorers and back-translation models for every row)}} \\
\rowcolor{refrow}Ground truth & 0 & 0 & 0 & 0 & 6.89 & 23.29 & 0 & 0 & 0 & 0 & 15.00 & 8.61 & 0 & 0 & 0 & 0 & 3.53 & 11.75 \\
SignSparK released$^\star$ & 10.26 & 2.06 & 8.46 & 11.16 & 0.00 & 0.13 & \multicolumn{6}{c}{\emph{not trained on PHOENIX}} & 10.39 & 2.44 & 8.77 & 12.06 & 0.56 & 0.72 \\
SignFlow, pose space & \snd{8.74} & 1.61 & \snd{7.35} & 9.42 & \snd{4.60} & 4.73 & 6.23 & 1.29 & 5.25 & 6.66 & \snd{15.07} & \textbf{7.12} & \snd{9.97} & \snd{2.27} & \textbf{8.49} & \snd{11.53} & \snd{1.68} & \snd{1.36} \\
SignFlow + Faur\'e et al.\ latent & \textbf{8.63} & \textbf{1.57} & \textbf{7.23} & \textbf{9.37} & 4.33 & \snd{5.22} & \textbf{6.02} & \snd{1.25} & \textbf{4.98} & \textbf{6.41} & 15.01 & \snd{6.86} & -- & -- & -- & -- & -- & -- \\
SignFlow + SOKE latent & 11.77 & 2.05 & 9.80 & 16.70 & 0.00 & 0.06 & 9.01 & 2.48 & 7.15 & 10.19 & 0.59 & 1.24 & -- & -- & -- & -- & -- & -- \\
\rowcolor{ourrow}\signtok\ + SignFlow (ours) & 8.85 & \snd{1.58} & 7.49 & \snd{9.41} & \textbf{5.99} & \textbf{8.14} & \snd{6.13} & \textbf{1.23} & \snd{5.16} & \snd{6.48} & \textbf{18.07} & 6.79 & \textbf{9.88} & \textbf{2.22} & \snd{8.54} & \textbf{11.07} & \textbf{2.87} & \textbf{2.93} \\
\bottomrule
\end{tabular}}
\caption{\textbf{Text-to-pose results.} DTW joint position error in the SOKE convention (accumulated cost in metres, lower is better), Procrustes-aligned (PA-JPE) and unaligned (JPE), for body and hands, and back-translation BLEU-4 (B-T) under two scorers: the model released by Low et al.~\cite{low2026signspark} (S) and Uni-Sign fine-tuned on the same poses (U). $\dagger$: as reproduced by Zuo et al.~\cite{zuo2025signs}; sources of all published rows are listed in the supplementary material. The two latent baselines are the same generator in the VAE latent of Faur\'e~\cite{faure2026impact} and in the pre-quantization latent of SOKE, trained with released code on the same poses; they were not run on How2Sign. $\star$: our run of the released checkpoint in text-only mode (retrieval off). The first two blocks are scored on other pose fits and listed for reference; the last block is a controlled comparison on identical clips, poses, and scorers. Bold: best, underlined: second best in the controlled block.}
\label{tab:main}
\end{table*}

\section{Experiments}
\label{sec:exp}

\paragraph{Datasets.}
We use three benchmark datasets, CSL-Daily \cite{zhou2021improving}, PHOENIX-2014T \cite{camgoz2018neural} and How2Sign \cite{duarte2021how2sign}, with the curated poses and splits released by Low et al.~\cite{low2026signspark}.

\paragraph{Implementation.}
Text features come from the frozen mBART-large-cc25 encoder~\cite{liu2020multilingual} applied to the input text. The tokenizer uses a masking ratio of 0.5, an 8-layer encoder ($d{=}256$), a 3-layer decoder and alignment windows of $k{=}8$ frames; the generator is SignFlow \cite{khan2025signflow} with 25 Euler steps. We compare four tokenizer variants: \emph{no text head} (reconstruction and kinematics only), \emph{+ sentence-level} (adding the sentence head), \emph{+ contextual tokens} (adding the alignment head with the contextual encoder states of mBART as targets), and \signtok\ (the alignment head with non-contextual input embeddings). Models are trained per corpus; all sizes, loss weights, and schedules are given in the supplementary material.

\paragraph{Metrics.}
Following SOKE \cite{zuo2025signs} and SignSparK \cite{low2026signspark}, we measure motion and spatial fidelity with DTW errors on Procrustes-aligned (DTW-PA-JPE) and unaligned (DTW-JPE) joint positions, and semantic fidelity is measured using back-translation BLEU-4 under two independent sign language translation models to avoid model bias: the model released with SignSparK (S) and Uni-Sign \cite{li2025uni} fine-tuned by us on the same poses (U); details are provided in the supplementary material. Table~\ref{tab:main} follows the SOKE convention (accumulated DTW cost in metres) for comparison with published numbers; the ablation reports the same errors per aligned frame in millimetres, together with the jerk of the generated motion relative to real signing. As human references, we report the \emph{ceiling}, the error between two signers producing the same sentence, and the \emph{floor}, the error between two signers producing different sentences.

\subsection{Results}
\label{sec:mainresults}

\paragraph{Against baselines (Table~\ref{tab:main}).}
In the controlled block, generating in the \signtok\ latent space improves on the released SignSparK model in its text-only mode (retrieval off) on every metric of CSL-Daily and How2Sign, by 11-23\% on the joint errors of CSL-Daily. Against the published text-only systems, \signtok\ has the lowest aligned hand-joint error on all three datasets and a higher body-joint error than the reported SignSparK result; these rows are scored on different pose fits, and SignSparK is a single multilingual model whereas ours are trained per corpus, so this comparison is indicative only.

\paragraph{Semantic fidelity.}
Back-translation separates the systems more than joint error does. Relative to pose-space generation, \signtok\ raises BLEU-4 under the S scorer by 20-70\% on all three datasets, and under the U scorer by 72\% on CSL-Daily and 115\% on How2Sign, while it is slightly lower on PHOENIX-2014T ($6.79$ against $7.12$). The released SignSparK model in text-only mode scores near zero under both scorers. This suggests that a semantically organized latent yields signing that is more often recognized as the intended sentence, even where it is not closer to the one reference. The two scorers disagree on PHOENIX-2014T, where S also rates generated clips above the real poses, so we read that corpus as inconclusive.

\subsection{Which latent is semantically organized?}
\label{sec:alignment}

\begin{table}[t]
\centering\scriptsize
\setlength{\tabcolsep}{3pt}
\renewcommand{\arraystretch}{1.08}
\begin{tabular}{l c cc}
\toprule
& & \multicolumn{2}{c}{Nearest neighbour} \\
\cmidrule(lr){3-4}
Representation & R@1$\uparrow$ & same sent.$\uparrow$ & same signer$\downarrow$ \\
\midrule
\rowcolor{grphead}\multicolumn{4}{c}{\textbf{No text}} \\
Raw pose & .061 & .023 & .955 \\
SOKE VQ-VAE (released code) & .060 & .090 & .779 \\
Fauré et al.\ VAE (released code) & .072 & .070 & .852 \\
Masked AE, no text head & .123 & .058 & .937 \\
\rowcolor{grphead}\multicolumn{4}{c}{\textbf{Text-aligned}} \\
\quad + sentence-level & .321 & .185 & .836 \\
\quad + contextual tokens & \snd{.355} & \snd{.349} & \snd{.711} \\
\rowcolor{ourrow}\signtok\ (word-level, ours) & \textbf{.569} & \textbf{.815} & \textbf{.377} \\
\bottomrule
\end{tabular}
\caption{\textbf{Organization of the latent}, CSL-Daily test split, mean-pooled clips, no tokenizer head involved. R@1: a ridge map fitted on the training split maps the sentence feature into the latent and retrieves the clip among 1{,}162. Nearest neighbour: fraction of clips whose cosine nearest neighbour is the same sentence signed by another person, or any clip by the same signer. The two external tokenizers are trained with their authors' code on the same poses. Bold: best, underlined: second best.}
\label{tab:align}
\end{table}

Table~\ref{tab:align} and Figure~\ref{fig:related-latent} answer the problem statement. Every representation learned without the text, ours and the two trained with released code, is nearly as signer-organized as the raw poses: a clip's nearest neighbour is the same signer in 78-96\% of cases and the same sentence by someone else in at most 9\%. Sentence-level alignment helps but leaves most of the signer structure in place, contextual tokens help more, and word-level alignment changes the latent qualitatively: the nearest neighbour is the same sentence by another person 82\% of the time, and same-sentence pairs lie at $0.08$ of the distance between random clips (Figure~\ref{fig:related-latent}). The same ordering holds on PHOENIX-2014T and, under a label-free word-sharing probe, on How2Sign (see the supplementary material).

\subsection{Does the head localize words?}
\label{sec:localization}
The contextual and the word-level heads differ in what they learn, not only in how much. We test a trained head with two sentences $A$, $B$ by the same signer sharing no content word: their latent windows are concatenated, and a word of $A$ is correct if its best-matching window lies in $A$ (chance $0.52$). A head that recognizes the sentence passes this with contextual tokens whether or not it knows the words; a head that localizes words passes it with non-contextual tokens. The contextual head scores $0.92$ with contextual tokens but $0.59$ with non-contextual ones: it has learned sentence identity, which every contextual token carries. \signtok's head scores $0.87$ with non-contextual tokens, picks the same word consistently across sentences and signers, and when the same sentence is signed by two people its word-to-window assignments agree above chance (supplementary material). This is consistent with Table~\ref{tab:align}: the head that localizes words is the one that organizes the latent by them.

\paragraph{Against other latents.}
The same generator in the reconstruction-trained VAE latent of Faur\'e et al.~\cite{faure2026impact} reaches slightly lower body errors than \signtok\ on CSL-Daily and PHOENIX-2014T (by at most $0.3$ in the units of Table~\ref{tab:main}), but its back-translation stays at the level of pose space ($4.33$ and $15.01$ against $5.99$ and $18.07$ for \signtok\ under S); the pre-quantization latent of SOKE is worse on every column. A latent trained for reconstruction thus keeps joint accuracy but brings no semantic gain, whereas the text-aligned latent does.

\subsection{Qualitative results}
\label{sec:qual}
\begin{figure*}[!ht]
\centering
\includegraphics[width=\textwidth]{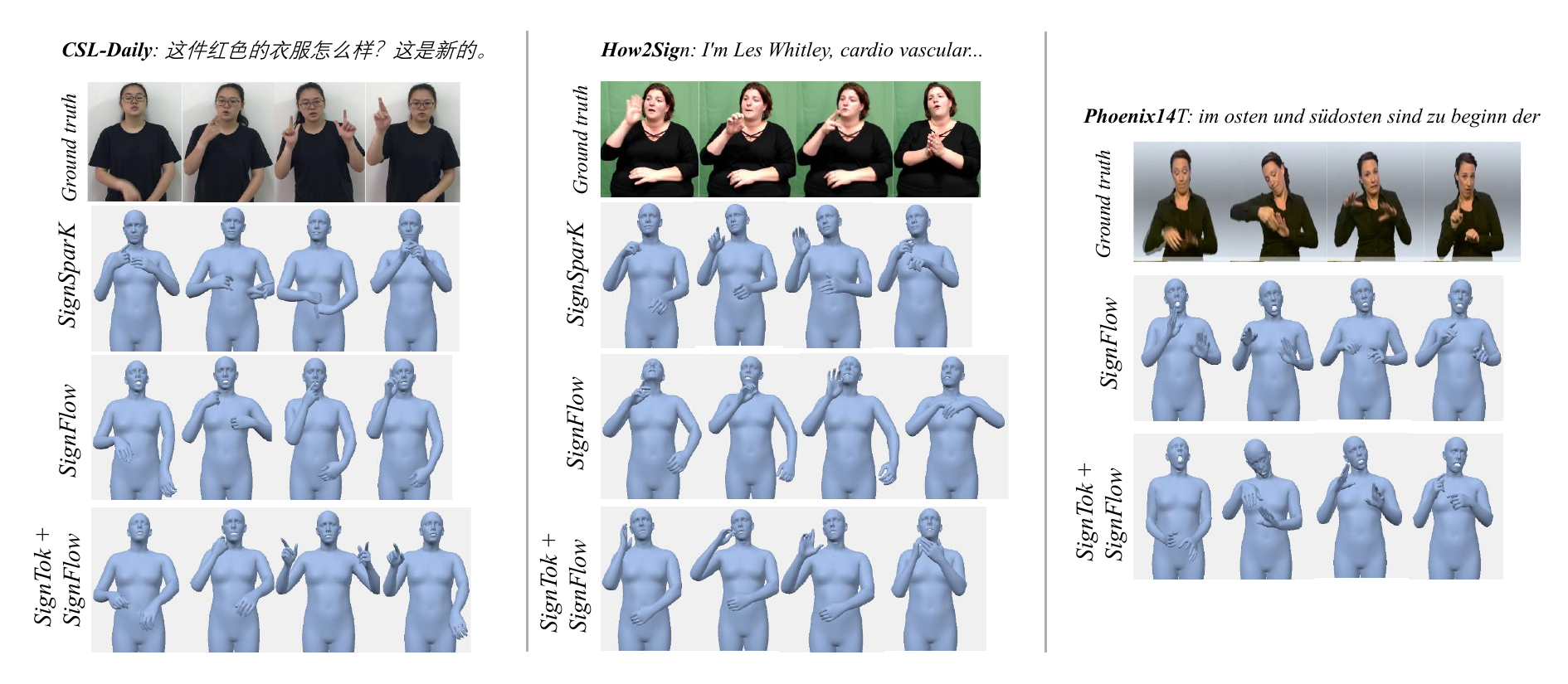}
\caption{\textbf{Generated signing on the three benchmarks.} One test clip per corpus, four evenly spaced frames per row: the reference video, the released SignSparK model (not available for PHOENIX-2014T, on which it was not trained), the pose-space generator SignFlow, and the same generator in the \signtok\ latent. Generated clips have their own timing, so frames at the same position do not show the same sign; what to compare is whether the handshapes of the sentence are formed and where the hands move.}
\label{fig:qual}
\end{figure*}

Figure~\ref{fig:qual} shows one clip per corpus; the pattern across clips is consistent with Table~\ref{tab:main}. Pose-space generation reaches the right region of the signing space but often misses the fine-grained finger configuration, and its motion is jittery between frames. Generation in the \signtok\ latent forms distinct handshapes, such as the index-finger and two-finger shapes of the CSL-Daily clip and the fist and flat hand of the PHOENIX clip, and moves smoothly between them. SignSparK in text-only mode produces fluent motion but tends to keep both hands low and close to the body, with less of the sentence's handshape vocabulary. On How2Sign, all systems are further from the reference, consistent with the larger errors and the weaker organization of every latent on that corpus (see supplementary material).

\section{Ablation}
\label{sec:ablation}

\begin{table}[!htb]
\centering\scriptsize
\setlength{\tabcolsep}{2.4pt}
\renewcommand{\arraystretch}{1.08}
\begin{tabular}{l cc cc c c}
\toprule
& \multicolumn{2}{c}{Body} & \multicolumn{2}{c}{Hand} & B-T$\uparrow$ & Jerk \\
\cmidrule(lr){2-3}\cmidrule(lr){4-5}
Generation space & PA$\downarrow$ & JPE$\downarrow$ & PA$\downarrow$ & JPE$\downarrow$ & B-4 & ratio \\
\midrule
\rowcolor{grphead}\multicolumn{7}{c}{\textbf{CSL-Daily}} \\
\rowcolor{refrow}Human, same sentence (ceiling) & 36.13 & 34.35 & 5.32 & 47.55 & 6.89 & 1.00 \\
Pose space (SignFlow) & \textbf{40.53} & \textbf{37.67} & 7.16 & 57.71 & 4.60 & 3.70 \\
Masked AE, no text head & 41.53 & 38.82 & 7.03 & 58.28 & 5.26 & 1.25 \\
\quad + sentence-level & \snd{40.70} & \snd{37.84} & \snd{6.80} & \snd{57.39} & \snd{5.53} & 1.27 \\
\quad + contextual tokens & 40.85 & 38.10 & \textbf{6.79} & \textbf{56.86} & 4.93 & 1.37 \\
\rowcolor{ourrow}\signtok\ (word-level, ours) & 41.19 & 38.32 & 6.97 & 57.85 & \textbf{5.99} & 1.49 \\
\rowcolor{refrow}Human, other sentence (floor) & 52.02 & 47.48 & 9.14 & 69.04 & -- & -- \\
\rowcolor{grphead}\multicolumn{7}{c}{\textbf{PHOENIX-2014T}} \\
\rowcolor{refrow}Ground truth & 0 & 0 & 0 & 0 & 15.00 & 1.00 \\
Pose space (SignFlow) & 36.73 & 33.09 & 7.74 & \snd{51.88} & 15.07 & 3.83 \\
Masked AE, no text head & 38.06 & 34.50 & 7.64 & 53.50 & 16.28 & 0.78 \\
\quad + sentence-level & \snd{36.51} & \textbf{32.74} & \snd{7.45} & 52.08 & 17.03 & 0.82 \\
\quad + contextual tokens & 37.23 & 33.40 & 7.47 & 52.39 & \snd{17.72} & 0.84 \\
\rowcolor{ourrow}\signtok\ (word-level, ours) & \textbf{36.49} & \snd{32.76} & \textbf{7.40} & \textbf{50.91} & \textbf{18.07} & 0.71 \\
\rowcolor{grphead}\multicolumn{7}{c}{\textbf{How2Sign}} \\
\rowcolor{refrow}Ground truth & 0 & 0 & 0 & 0 & 3.53 & 1.00 \\
Pose space (SignFlow) & \snd{41.33} & \snd{37.70} & 9.00 & 60.75 & 1.68 & 1.26 \\
Masked AE + sentence-level & 42.51 & 38.97 & \snd{8.98} & \snd{60.24} & \snd{2.73} & 0.85 \\
\rowcolor{ourrow}\signtok\ (word-level, ours) & \textbf{40.22} & \textbf{36.87} & \textbf{8.69} & \textbf{57.99} & \textbf{2.87} & 0.79 \\
\bottomrule
\end{tabular}
\caption{\textbf{Ablation of the generation space.} Per-frame DTW error in mm (lower is better), back-translation BLEU-4 (S scorer), and jerk relative to real signing ($1$ = natural, $>1$ jittery, $<1$ over-smoothed). All tokenizer variants keep the masked autoencoder and the kinematics head. Single runs; seed-to-seed noise is $0.2$-$0.5$\,mm on body. Bold: best, underlined: second best.}
\label{tab:ablation}
\end{table}

\paragraph{Latent against pose space.}
The cleanest comparison is with the same generator in pose space, since the two systems differ only in the space they generate in. On the per-frame errors of Table~\ref{tab:ablation}, \signtok\ is better on every column on How2Sign, better on the hands and level on the body on PHOENIX-2014T, and, on CSL-Daily, better on the aligned hand error and $0.7$\,mm (under $2\%$) behind on the body. Joint accuracy is therefore comparable between the two spaces. The clearer difference is in the motion itself: pose-space generation is jittery, with a jerk $3.7$-$3.8$ times that of real signing on CSL-Daily and PHOENIX-2014T, against $0.7$-$1.5$ in the latent. Both systems remain above the human ceiling on CSL-Daily, by $14\%$ on the body and $31\%$ on the hands for \signtok, so the hands carry the larger remaining gap.

\paragraph{Text alignment is what makes the latent generable.}
The masked autoencoder without a text head has the highest body error of all rows, $1.0$-$1.3$\,mm behind pose space on the two corpora where it was run, which is above seed noise. Adding any text head lowers the body error and raises back-translation on PHOENIX-2014T, and the sentence-level head does both on CSL-Daily. Without a text head, our masked autoencoder is therefore a worse generation space than the poses themselves, and text alignment is what recovers and improves on it.

\paragraph{Alignment granularity.}
On joint error, the three text heads generate within about $0.5$\,mm of each other on CSL-Daily, where the sentence-level head has the lowest body error; on PHOENIX-2014T the word-level head is the best or level row on every column, and on How2Sign it is the best row on every column. On back-translation, the word-level head is the best row on every corpus: the ordering is no text head $<$ sentence-level $<$ word-level on CSL-Daily and PHOENIX-2014T, and sentence-level $<$ word-level on How2Sign, with the contextual-token head between the two on PHOENIX-2014T and the lowest row on CSL-Daily. Finer alignment therefore brings semantic organization (Table~\ref{tab:align}) and semantic fidelity on every corpus, and joint accuracy on How2Sign. Reconstruction errors, second seeds, and ablations of the other two heads, the masking ratio and guidance are given in the supplementary material.

\section{Conclusion}
\label{sec:conclusion}

This paper presents \signtok, a continuous pose tokenizer for text-conditioned sign language generation. We show that pose space and reconstruction-trained tokenizers are organized largely by signer identity rather than linguistic content: the nearest neighbor of a clip is the same sentence by another signer in at most 9\% of cases on CSL-Daily. \signtok\ shapes its latent with three training-only heads for kinematics, sentence-level alignment, and word-level alignment between latent windows and content words, which raises that rate to 82\%, and we show that the word-level head localizes individual words rather than recognizing sentences. On three benchmark datasets, the same generator trained in the \signtok\ latent raises back-translation BLEU-4 over pose space in five of six benchmark-scorer settings at comparable joint accuracy, with smoother motion. The latent does not lower joint error consistently; the signer remains encoded in it. These are directions for future work. Our results indicate that organizing a continuous pose representation by the words of the sentence, rather than for reconstruction alone, makes it a better space for text-conditioned generation.

\bibliography{reference.bib}
\bibliographystyle{ieeetr}

\end{document}